\documentclass[sn-mathphys]{sn-jnl}

\jyear{2022}%

\theoremstyle{thmstyleone}%
\theoremstyle{thmstyletwo}%

\newtheorem{assumption}{Assumption}[section]

\theoremstyle{thmstylethree}%

\usepackage{graphicx}
\DeclareGraphicsExtensions{.pdf,.jpeg,.png}
\usepackage{epstopdf}
\usepackage{txfonts}
\usepackage{subfigure}
\usepackage[graphicx]{realboxes}

\usepackage{algorithm}
\usepackage{algorithmicx}
\usepackage{algpseudocode}
\usepackage{array}
\usepackage{amsmath}
\usepackage{amsthm}
\usepackage{amsfonts}
\usepackage{amssymb}
\usepackage{cancel}
\usepackage{multirow} 
\usepackage{xcolor}
\usepackage{setspace}
\makeatletter
\newcommand{\removelatexerror}{\let\@latex@error\@gobble}
\makeatother
\usepackage{booktabs}
\usepackage{url}
\usepackage{footnote}
\usepackage{threeparttable}
\allowdisplaybreaks[4]
\usepackage[figuresright]{rotating}

\begin{document}

\title[Article Title]{Learning Diverse Policies with Soft Self-Generated Guidance}


\author[1,2]{\fnm{Guojian} \sur{Wang}}\email{wgj@buaa.edu.cn}

\author[2,3,4]{\fnm{Faguo} \sur{Wu}}\email{faguo@buaa.edu.cn}

\author*[1,2,4]{\fnm{Xiao} \sur{Zhang}}\email{xiao.zh@buaa.edu.cn}

\author[1,2]{\fnm{Jianxiang} \sur{Liu}}\email{sy2109127@buaa.edu.cn}

\affil*[1]{\orgdiv{School of Mathematical Sciences}, \orgname{Beihang University}, \orgaddress{\city{Beijing}, \country{China}}}

\affil[2]{\orgdiv{Key Laboratory of Mathematics, Informatics and Behavioral Semantics}, \orgname{Education of Ministry}, \orgaddress{\city{Beijing}, \country{China}}}


\affil[3]{\orgdiv{Institute of Artificial Intelligence}, \orgname{Beihang University}, \orgaddress{\city{Beijing}, \country{China}}}

\affil[4]{\orgdiv{Zhongguancun Laboratory}, \orgaddress{\city{Beijing}, \country{China}}}


\abstract{
  Reinforcement learning (RL) with sparse and deceptive rewards is challenging because non-zero rewards are rarely obtained. Hence, the gradient calculated by the agent can be stochastic and without valid information. Recent studies that utilize memory buffers of previous experiences can lead to a more efficient learning process. However, existing methods often require these experiences to be successful and may overly exploit them, which can cause the agent to adopt suboptimal behaviors. This paper develops an approach that uses diverse past trajectories for faster and more efficient online RL, even if these trajectories are suboptimal or not highly rewarded. The proposed algorithm combines a policy improvement step with an additional exploration step using offline demonstration data. The main contribution of this paper is that by regarding diverse past trajectories as guidance, instead of imitating them, our method directs its policy to follow and expand past trajectories while still being able to learn without rewards and approach optimality. Furthermore, a novel diversity measurement is introduced to maintain the team's diversity and regulate exploration. The proposed algorithm is evaluated on discrete and continuous control tasks with sparse and deceptive rewards. Compared with the existing RL methods, the experimental results indicate that our proposed algorithm is significantly better than the baseline methods regarding diverse exploration and avoiding local optima.
}




\maketitle

\section{Introduction}
In recent years, deep reinforcement learning has been demonstrated to effectively solve sequential decision-making problems in many application domains, such as computer and board games playing~\cite{Mnih2015HumanlevelCT, Silver2016MasteringTG}, continuous control~\cite{Lillicrap2016ContinuousCW, Schulman2015TrustRP,fujimoto2018addressing}, and robot navigation~\cite{nair2018overcoming}. Despite these success stories, reinforcement learning with sparse and deceptive rewards remains a challenging problem in the field of RL~\cite{houthooft2016vime, Florensa2017StochasticNN,oh2018self} because making a good trade-off between exploration and exploitation becomes more intractable in tasks with sparse and deceptive rewards.

Optimizing with sparse feedback requires agents to reproduce past good trajectories efficiently and avoid being struck in local optima. In tasks with large state spaces and sparse rewards, a desired positive reward can only be received after the agent continuously executes many appropriate actions. Hence, the agent can rarely collect highly rewarded trajectories. Meanwhile, the gradient-based parameter update of modern deep RL algorithms might result in a catastrophic forgetting of past experiences because the gradient-based parameter update is incremental and slow, and it has a global impact on all parameters of the policy and the value function~\cite{guo2020memory}. Therefore, the agent might suffer from severe performance degradation when the ideal trajectories with the highest returns are rarely collected, incurring an unstable policy optimization process. Finally, agents can adopt sub-optimal myopic behaviors and be struck in local optima due to overly exploiting past imperfect experiences and not exploring the state-action space systematically.

Such tasks with sparse and deceptive reward signals are common in real-world problems. Recently, much work has studied how using the non-parametric memory of past experiences improves policy learning in RL. Prioritized experience replay~\cite{schaul2015prioritized} proposes prioritizing past experiences before learning the policy parameters from them. Self-imitation learning~\cite{gangwani2019learning,oh2018self,guo2018generative} builds a memory buffer to store past good trajectories and thus can rapidly learn the near-optimal strategy from these past experiences when faced with a similar situation. Memory-augmented policy optimization~\cite{liang2018memory} leverages a memory buffer of prior highly rewarded trajectories to reduce the estimated variance of the policy gradient. Episodic reinforcement learning~\cite{lin2018episodic} uses past good experiences stored in an episodic memory buffer to supervise an agent and force the agent to learn good strategies. Model-free episodic control~\cite{blundell2016model} and neural episodic control~\cite{pritzel2017neural} use episodic memory modules to estimate the state-action values. Diverse trajectory-conditioned self-imitation learning~\cite{guo2020memory} proposes learning a novel trajectory-conditioned policy that follows and expands diverse trajectories in the memory buffer.

This existing work uses non-parametric memories of past good experiences to rapidly latch onto successful strategies and improve the learning efficiency of policy and value function. However, we must note that exploiting the past good trajectories described in the aforementioned work might hurt the agent's performance in tasks with sparse and deceptive reward functions. Two main reasons can lead to the performance degradation of algorithms in tasks with sparse and deceptive rewards. First, the past self-generated trajectories stored in the memory buffer are imperfect. The trajectories in the memory buffer are not gold trajectories but highly rewarded trajectories collected by accident. Second, the RL agent limits its exploration to a small portion of the state action space because of prior experience and network initialization~\cite{peng2020non}. This way, the agent can quickly generate trajectories leading to sub-optimal goals. Exploiting these successful suboptimal trajectories with limited directions might cause the agent to learn myopic behaviors. This will limit the agent's exploration region and prevent the agent from discovering alternative strategies with higher returns.

In this paper, we conduct a practical RL algorithm by regarding previous diverse trajectories as guidance in the sparse reward setting, even if these trajectories are suboptimal or not highly rewarded. Our critical insight is that we can utilize imperfect trajectories with or without sparse rewards to regulate the direction of policy optimization while preserving the diversity of agents by virtue of two steps. In the first policy improvement (PI) step, we develop a new method that exploits the self-generated guidance to enable the agent to reproduce diverse past trajectories efficiently while encouraging agents to expand these trajectories smoothly and gradually visit underexplored regions of the state action space. Specifically, our method guides agents to revisit the regions where past good trajectories are located by minimizing the distance of state representations of trajectories. Meanwhile, our method allows for flexibility in the action choices to help the agent choose different actions and visit novel states. In the second policy exploration (PE) step, we introduce a novel diversity measurement to drive the different agents on the team to reach diverse regions in the state space and maintain the diversity of an ensemble of agents. By designing this new diversity measurement, our algorithm does not have to maintain a set of auto-encoders~\cite{zhang2019learning} and can prevent the agents from being stuck in local optima. Our main contributions are summarized as follows:
\begin{enumerate}
\item We develop a novel two-step RL framework that makes better use of diverse self-generated demonstrations to promote learning performance in tasks with sparse and deceptive rewards.
\item To the best of our knowledge, this is the first study that regards self-generated imperfect demonstration data as guidance and shows the importance of exploiting these previous experiences to drive exploration indirectly.
\item We illustrate that by using the agent's self-generated demonstration trajectories as guidance, the agent can reproduce diverse past trajectories quickly and then smoothly move beyond to result in a more effective policy. 
\item A new diversity metric for the ensemble of agents has been proposed to achieve deep exploration and avoid being stuck in local optima.
\item Our method achieves superior performance over other state-of-the-art RL algorithms on several challenging physical control benchmarks with sparse and deceptive rewards in terms of diverse exploration and improving learning efficiency.
\end{enumerate}

The rest of this article is organized as follows. Section~\ref{sec:related} describes the progress of the related work. Section~\ref{sec:backgroung} briefly describes the preliminary knowledge for the article. Section~\ref{sec:approach} introduces our proposed method for reinforcement learning with sparse and deceptive rewards. Experimental results are presented in Section~\ref{sec:experience}. Finally, we draw our conclusions in Section~\ref{sec:conclusion}.

\section{Related Work}\label{sec:related}
\textbf{Exploration and Exploitation}
It is a long-standing and intractable challenge to balance exploration and exploitation in the field of RL. The exploration enables the agent to visit the under-discovered state-action space and collect trajectories with higher returns. On the contrary, exploitation encourages the agent to use what it already knows to maximize the expected returns. Many work aims to improve the exploration ability of the RL agent. Some work proposes to add stochastic noise to the output actions~\cite{Mnih2015HumanlevelCT,Schulman2015TrustRP,schulman2017proximal,Lillicrap2016ContinuousCW,fujimoto2018addressing}, or parameters of policy and value networks~\cite{plappert2017parameter,fortunato2019noisy,osband2019deep} to encourage exploration. Many other work defines the concept of intrinsic reward to promote the agent to visit the under-explored state-action space~\cite{pathak2017curiosity,achiam2017surprise,savinov2018episodic}. Furthermore, plenty of studies introduce a new optimization objective to change the gradient update direction of parameters~\cite{hong2018diversity,han2020diversity,peng2020non,masood2019diversity}. Another straightforward idea to expand the agent's exploratory area is to employ a team of agents to explore the environment collaboratively and share the collected experiences with each other~\cite{mnih2016asynchronous,schmitt2020off,espeholt2018impala}. In all these methods, although the agent can access the under-discovered area due to randomness and artificial incentive, it is still difficult for the agent to get rid of local optima in long-horizon, sparse reward tasks because it is rare for the agent to collect trajectories with non-zero rewards in these hard-exploration environments. Our method maintains different memory buffers to store past good trajectories for each agent in the team, and an agent only shares past good trajectories with each other to calculate a diversity measurement.

\textbf{Memory Based RL} 
A memory buffer enables the agent to store and utilize past experiences to aid online RL training. Many prior work proposed storing past good experiences in replay buffers with a prioritized replay mechanism to accelerate the training process~\cite{schaul2015prioritized,hester2018deep,nair2018overcoming}. Episodic reinforcement learning methods~\cite{pritzel2017neural,blundell2016model,lin2018episodic} memorize past good episodic experiences by maintaining and updating a look-up table and act upon these good experiences in the decision-making process. Self-imitation learning (SIL) methods~\cite{oh2018self,gangwani2019learning} train the agent to imitate the highly rewarded trajectories with the SIL and GAIL objectives, respectively. There are many previous works where the agent learns diverse exploratory policies based on episodic memory~\cite{guo2020memory,badia2020never}. Unlike these aforementioned methods, our method encourages the agent to visit the part of the state space where the agent can obtain higher rewards by calculating the distance between the current trajectory and the past good trajectory.

\textbf{Imitation Learning (IL)}
Imitation learning aims to train a policy by imitating the demonstration data generated by human experts. Behavior Clone (BC) is a simple IL approach where the unknown expert policy is estimated from demonstration data by supervised learning. BC methods, however, usually suffer from the heavy distribution shift problem~\cite{ross2011reduction}. Inverse reinforcement learning (IRL) solves forward RL problems by recovering the reward function from demonstration data~\cite{Ng2000AlgorithmsFI,ziebart2008maximum}. Generative Adversarial Imitation Learning (GAIL)~\cite{Ho2016GenerativeAI} formulated the IL problem as a distribution matching problem, which can avoid estimating the reward function. All these IL methods rely on the availability of high-quality and sufficient human demonstrations. In contrast, our method treats the past good trajectories generated by the agent as demonstration data.

\section{Preliminaries}\label{sec:backgroung}
This study shows that the MMD metric can be used as a distance constraint to prevent the agent from falling into local optima. Using the exterior penalty function method, we transform the constrained RL optimization problem into an unconstrained optimization problem, and the MMD distance can be used to define an intrinsic reward. Our method can be naturally combined with the hierarchical reinforcement learning algorithm, and the policy gradient can adjust pre-trained skills and the high-level policy during the training phase.

\subsection{Reinforcement Learning} 
We consider a discounted Markov decision process (MDP) defined by a tuple $ M = (\mathcal{S}, \mathcal{A}, P, r, \rho_{0}, \gamma) $, in which $ \mathcal{S} $ is a continuous state space, $ \mathcal{A} $ is a (discrete or continuous) action space, $ P: \mathcal{S} \times \mathcal{A} \times \mathcal{S} \rightarrow \mathbb{R}_{+} $ is the transition probability distribution, $ r: \mathcal{S} \times \mathcal{A} \rightarrow [R_{min}, R_{max}] $ is the reward function, $ \rho_{0}: \mathcal{S} \rightarrow \mathbb{R}_{+}$ is the distribution of the initial state $ s_{0} $, $ \gamma \in [0,1] $ is a discount factor. A stochastic policy $ \pi_{\theta} : \mathcal{S} \rightarrow \mathcal{P} (\mathcal{A})$ parameterized by $ \theta $, maps the state space $ \mathcal{S} $ to the set of probability distributions $\mathcal{P}(\mathcal{A})$ over the action space $\mathcal{A}$. The state-action value function is defined as $Q(s_t,a_t)=\mathbb{E}_{\rho_0, P, \pi}\left[\sum_{t^{\prime}=t}^{\infty}\gamma^{t^{\prime}-t} r(s_{t^{\prime}},a_{t^{\prime}})\right]$.

In general, the objective of RL algorithms is to find an optimal policy $\pi_{\theta}$ that maximizes the expected discounted return:
\begin{equation}
  \eta(\pi_{\theta}) = \mathbb{E}_{\tau}\left[\sum_{t=0}^{\infty}\gamma^{t}r(s_{t},a_t)\right],
\end{equation}
We use $ \tau = (s_0,a_0,\dots) $ to denote the entire history of the state, action pairs, where $s_0\sim \rho_0(s_0)$, $a_t\sim\pi_{\theta}(a_t \vert s_t)$, and $s_{t+1}\sim P(s_{t+1} \vert s_t,a_t) $.

Similar to~\cite{sutton1999policy}, when $\gamma=1$, we define the stationary state visitation distribution for the policy $\pi_\theta$ by $\rho_{\pi}(s)=\lim_{t\rightarrow\infty}P(s_t=s \vert s_0,\pi)$, where the initial state $s_{0} \sim \rho_0$. The expected discounted return can be rewritten as $\mathbb{E}_{\rho_{\pi}(s,a)}\left[r(s,a)\right]$, where $\rho_{\pi}(s,a)=\rho_{\pi}(s)\pi_{\theta} (a \vert s)$ is the state-action visitation distribution.

\subsection{Policy Gradient Algorithms}
This paper's study is based on policy gradient RL algorithms that directly compute policy gradients to optimize the agent's policy parameters. We briefly introduce a well-known policy gradient algorithm: Proximal Policy Optimization (PPO).

\subsubsection{PPO}
PPO~\cite{schulman2017proximal} uses a clipped objective function to constrain the step size during an update and prevent drastic parameter changes. This leads to a stable training process compared with other algorithms. The PPO agent adjusts the parameters of the policy by maximizing the following clipped objective function:
\begin{equation*}
  \max_{\theta}\mathcal{L}(\theta)=\mathbb{E}_{\theta}\left[\min\left(
    \frac{\pi_\theta(a_t \vert s_t)}{\pi_{\theta_{old}}(a_t\vert s_t)}, {\rm clip}\left(
      \frac{\pi_\theta(a_t\vert s_t)}{\pi_{\theta_{old}}(a_t\vert s_t)}, 1-\epsilon, 
      1+\epsilon\right)\right)A(s_t,a_t)\right],
\end{equation*}
where $\pi_\theta(\cdot)$ and $\pi_{\theta_{old}}(\cdot)$ represent the current policy and the old policy, respectively, and $\epsilon$ is the clipping ratio which is a hyperparameter that is empirically determined.

\subsection{Maximum Mean Discrepancy}
Maximum Mean Discrepancy (MMD) can be used to measure the difference (or similarity) between two probability distributions~\cite{gretton2006kernel, Gretton2012OptimalKC,masood2019diversity,wang2023adaptive}. Let $ \boldsymbol{x}:= \{ x_0, x_1, \cdots, x_l \} $, and $ \boldsymbol{y}:= \{ y_0, y_1, \cdots, y_m \} $ be two sets of samples which are taken independently and identically from two distributions $ p $ and $ q $.$ p $ and $ q $ are defined in a nonempty compact metric space $ \mathbb{X} $. Then, we can define MMD as:
\begin{equation}
  \label{equ:MMD}
  {\rm MMD}(p, q, \mathcal{F}) := \sup_{f \in \mathcal{F}} 
  \left(\mathbb{E}_{x \sim  p}\left[f({x})\right] - 
  \mathbb{E}_{y \sim q}\left[f({y})\right]\right)
\end{equation}
where $x\in\boldsymbol{x}, y\in\boldsymbol{y}$, and $\mathcal{F}$ a class of functions on $\mathbb{X}$. If $\mathcal{F} $ satisfies the condition $ p = q $ if and only if $ \mathbb{E}_{{x} \sim p}[f({x})] = \mathbb{E}_{{y} \sim q}[f({y})], \forall f \in \mathcal{F}$, then MMD is a metric in the space of probability distributions in $\mathbb{X}$ measuring the discrepancy between any two distributions $ p $ and $ q $~\cite{fortet1953convergence}.

What kind of function class makes the MMD a metric? According to~\cite{gretton2006kernel}, the space of bounded continuous functions $\mathcal{F}$ on $\mathbb{X}$ satisfies the condition, but it is intractable to compute the MMD with finite samples in such a considerable function class. Fortunately, when $ \mathcal{F} $ is a reproducing kernel Hilbert space $\mathcal{H}$ defined by kernel $ k(\cdot, \cdot) $, it is enough to uniquely identify whether $ p = q $ or not and the MMD is tractable in the space:
\begin{equation*}
  \begin{aligned}
    {\rm MMD}^2 (p, q, \mathcal{H}) = \mathbb{E}
    [k({x}, {x}^\prime)]
    - 2 \mathbb{E}
    [k({x}, {y})] 
    + \mathbb{E}[k({y}, {y}^\prime)]
  \end{aligned}
\end{equation*}

\section{Proposed Approach}\label{sec:approach}
In this section, we formulate a novel RL framework named Policy Optimization with Soft self-generated guidance and diverse Exploration (POSE). The proposed method utilizes a team of agents to explore the environment simultaneously and encourages them to visit non-overlapping areas of state spaces. Every agent in the team maintains a memory buffer storing past good trajectories, and these offline data can be regarded as guidance to enable the agents to revisit diverse regions in the state space where the agents can receive high rewards and drive deep exploration.

\subsection{Overview of POSE}
\label{sec:overview}
One feasible approach to better explore challenging tasks with sparse and deceptive rewards is employing a team of agents and forcing them to explore different parts of the state-action space. In this way, diverse policies can be learned by different agents in the team, which prevents agents from being stuck in local optima. 

\begin{figure}[htb]
  \centering
  \includegraphics[scale=0.415]{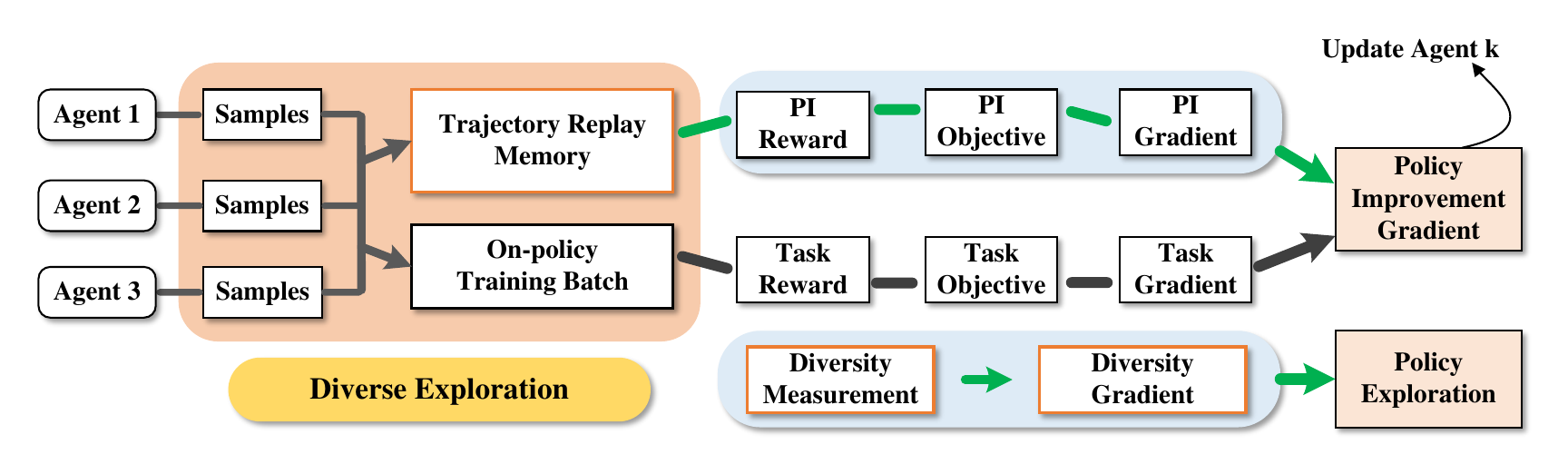}
  \caption{
    The framework of POSE. The diverse exploration in POSE leverages multiple agents to sample training batches and use the measurement of diversity to encourage agents to collect diverse trajectories. In contrast, the traditional RL, e.g., PPO~\cite{schulman2017proximal} or SAC~\cite{haarnoja2018soft}, uses a single agent to collect data. In the meantime, every agent maintains a replay buffer and stores specific trajectories in this buffer. These past good trajectories can guide the agents to revisit the region where the agents can obtain rewards with a higher probability.
  }
  \label{fig:diverse_exploration}
\end{figure}

As shown in Fig.~\ref{fig:diverse_exploration}, our POSE method employs a team of $K$ agents to interact with the environment and generate many state-action sequences. Different from the multi-agent RL setting~\cite{lowe2017multi}, where all agents live in a shared environment, and the action of an agent can affect other agents' states and action choices, in our design, each agent exists in an independent copy of the same environment and has no interaction with other agents in the team when sampling data. Every team agent collects an on-policy training batch $\mathcal{B}$ containing $M$ trajectories in the environment in each training iteration. Meanwhile, we also maintain a replay buffer $\mathcal{M}$ for each agent to store specific trajectories generated in previous rollouts. Further, POSE is presented in Algorithm~\ref{algo:DS2L} of Appendix~\ref{appendix:A} in detail. Next, We will explain how to organize the trajectory buffer.

\textbf{Organizing Trajectory Replay Memory for Exploration and Exploitation}\quad 
We maintain a trajectory replay memory $\mathcal{M}^i=\{(\tau_{(1)},e_{(1)},n_{(1)}),(\tau_{(2)},e_{(2)}, n_{(2)}),\cdots\}$ for the $i$-th agent of the team. The number of trajectories in $\mathcal{M}^i$ is no more than $k$, and hence $\tau_{(j)}$ in $\mathcal{M}^i$ is one of the top-$k$ trajectories ending with the state embedding $e_{(j)}$. $n_{(j)}$ is the number of steps of the trajectory $\tau_{(j)}$. Different from the SIL method~\cite{oh2018self}, in which only the successful trajectory with a return above a certain threshold is eligible to be stored in the replay memory, the imperfect trajectories that are not highly rewarded or even do not reach any goal can be considered as offline demonstrations in our method. For example, the trajectory is on the path to some goal, although it does not reach the goal. Furthermore, every replay memory only stores trajectories with a similar embedding that corresponds to the same goal or state region. If the embedding $e$ of a new trajectory $\tau$ is similar to the trajectories in the $i$-th agent's memory and this trajectory is better than (i.e., higher return, shorter trajectory length or shorter distance to the goal) the worst trajectory of this memory, we replace this worst trajectory by this new entry $\{\tau,e,n\}$.

\textbf{Guiding Agent to Reproduce Trajectory to State of Interest}\quad
To achieve better performance than existing RL methods in the sparse and episodic reward setting, we introduce a novel method that is beneficial to improve the data efficiency of RL algorithms and help the agent reproduce previous trajectories in the memory buffer efficiently. We introduce a new distance measure that calculates the difference between different trajectories and then develop an RL optimization problem with distance constraints by regrading diverse past demonstrations as guidance. Intuitively, our method can encourage the agent to revisit the parts of state space where the past trajectories are located. We will introduce how to train the policy in detail in Sec.~\ref{sec: soft self-imitation learning}

\textbf{Improving Exploration by Generating Diverse Trajectories}\quad
Compared with typical distributed RL methods like A3C~\cite{mnih2016asynchronous} and IMPALA~\cite{espeholt2018impala}, our POSE method not only simply employs multiple agents to collect amounts of trajectories independently by interacting with the parallel environments but also uses a diverse exploration mechanism to ensure exploration efficiency. We propose a new diversity metric to drive the different agents on the team to reach diverse regions in the state space and maintain the diversity of an ensemble of agents. An agent in the team could only pay more attention to visiting the state underexplored by other agents. Consequently, this helps the agents in the team to explore the environment systematically and avoid being stuck in local optima.

\subsection{Policy Improvement with Soft Self-Generated Guidance}
\label{sec: soft self-imitation learning}
We assign a trajectory $\tau$, a domain-dependent behavior characterization $b(\tau)$ to describe its behavior. For example, in MuJoCo Maze tasks described as the benchmark~\cite{duan2016benchmarking}, $b(\tau)$ can be as simple as a two-dimensional vector sequence, and components in the sequence represent the agent's $(x,y)$ location in every time-step:
\begin{equation}
  b(\tau) = b(\{(s_0,a_0),\cdots,(s_n,a_n)\})=\{(x_0,y_0), \cdots, (x_n,y_n)\}.
\end{equation}
The behavior characterization $b(\tau)$ suggests that states or position information, not actions, are used to distinguish different trajectories, and a similar approach is adopted by~\cite{conti2018improving,eysenbach2018diversity}. Other behavior characterization functions can also be defined to adjust the focus of the distance measurement based on different aspects such as state visits, action choices, or both.

A particular distance between the current trajectory $\tau$ and the replay memory $\mathcal{M}$ can be computed as follows:
\begin{equation}
  \label{equ:dist}
  \operatorname{dist}(\tau,\mu)=D_{\rm MMD}(b(\tau),b(\mu))={\rm MMD}^2(p,q,\mathcal{H}).
\end{equation}
Here, $b(\cdot)$ extracts the coordinate information corresponding to each state-action pair of the trajectory. Meanwhile, $\tau$ and $\mu$ are viewed as two deterministic policies in Eq.~\eqref{equ:dist}, and $p$ and $q$ are state-action distributions that are induced by deterministic policies $\tau$ and $\mu$, respectively. Furthermore, the distance between the current trajectory $\tau$ and the trajectory replay memory $\mathcal{M}$ is defined as follows:
\begin{equation}
  \operatorname{dist}(\tau,\mathcal{M}) = \min_{\mu\in\mathcal{M}}D_{\rm MMD}(b(\tau), b(\mu)).
\end{equation}
i.e. $\operatorname{dist}(\tau,\mathcal{M})$ is the minimum of distances between $\tau$ and 
each trajectory in the replay memory $\mathcal{M}$.

Existing methods of maintaining a memory buffer may overly exploit those good experience data, but these trajectories collected during the training process can be imperfect. Excessive exploitation of imperfect demonstrations might lead to myopic behaviors and sometimes hurt performance. We update the agent's parameters by regarding previous good trajectories as guidance rather than directly imitating these imperfect trajectory data. By imposing distance constraints in the trajectory space, each team agent is encouraged to revisit the region where past good trajectories are located. This way, the agent exploits what it already knows to maximize reward and reduces the overuse of previous good trajectories. Furthermore, our method allows for flexibility in the action choices and enables the agent to smoothly move beyond to find near-optimal policies. This intuition is based on the following observation:
\begin{assumption}
  \label{assu:1}
  A bounded tolerance factor $d$ exists such that trajectories with higher returns always stay within an area closer to the demonstration trajectories specified by $d$, even when the demonstration trajectories are imperfect and generated by the agent interacting with the environment. 
  \begin{equation}
    \forall \mu \in \mathcal{M}, \exists d \in (0, \infty), 
    \operatorname{dist}(\tau, \mu) \le d, R(\tau) \ge R(\mu),
  \end{equation}
\end{assumption}

Based on this distance in the trajectory space, we define a new RL optimization problem with constraints for the $i$-th agent as follows:
\begin{equation} 
  \label{equ:dist_constraint}
  \begin{aligned}
    \theta_{k+1}^{i}&=\max_{\theta^{i}} \, J(\theta^{i}_{k}), \\
    {\rm s.t.}\; \operatorname{dist}(\tau_{(j)}&,\mathcal{M}^i)\le\delta,\forall\tau_{(j)}
    \in \mathcal{B}^{i}.
  \end{aligned}
\end{equation} 
Here $\delta$ is a constant, and $\mathcal{M}^i$ is the replay memory of the $i$-th agent. $\mathcal{B}^i$ contains the trajectory data collected by the $i$-th agent at the current epoch and $\tau_{(j)}$ represents a trajectory in the buffer $\mathcal{B}^i$. From the perspective of policy optimization, it indicates that using constraints would better fit our problem settings for two reasons. 

\textbf{1. Convergence.} 
The constraint could affect the policy update when there are trajectories that do not satisfy the constraints. In this manner, it directs the agent to generate trajectories that stay in the constraint set defined by the distance on the trajectory space. According to Assumption~\ref{assu:1}, the more frequently the agent visits the state space around demonstration trajectories, the more likely the agent is to produce trajectories with higher returns.

\textbf{2. Optimality.} 
The agent's replay buffer containing previous specific trajectories is maintained by a dynamic update mechanism. Once the agent collects better trajectories with higher returns, shorter trajectory length, or shorter distance to the goal, the worst trajectories in the buffer will be replaced. Consequently, compared with the self-imitation learning methods, our method can leverage the imperfect demonstration trajectories for guiding the policy while eliminating their side effects in optimization, thus working better with imperfect demonstration trajectories.

\textbf{Optimization Process of Soft Self-Imitation}\quad
This section mainly describes how to efficiently optimize the RL objective with the distance constraint on the trajectory space. To simplify, we have omitted the superscript $i$ for some symbols, and these symbols have the same meaning as above unless otherwise specified. Firstly, using the Lagrange multiplier method, the optimization problem~\eqref{equ:dist_constraint} can be converted into an unconstrained form:
\begin{equation}
\label{equ: P_I}
  \begin{aligned}
    L(\theta) = J(\theta)
    - \sigma\mathbb{E}_{\tau\in\mathcal{B}^{i}}\operatorname{d}(\tau,\mathcal{M}^{i}),
  \end{aligned}
\end{equation}
where $\sigma > 0$ is a Lagrange multiplier which is used to determine the effect of the constraint, and $\operatorname{d}(\tau,\mathcal{M}^{i}) = 0$, if $\operatorname{dist}(\tau,\mathcal{M}^{i}) \le \delta$, else $\operatorname{d}(\tau,\mathcal{M}^{i}) = \operatorname{dist}(\tau,\mathcal{M}^{i})$. Then, the gradient with respect to the policy parameters $\theta$ of the objective~\eqref{equ: P_I} is expressed as:
\begin{equation}
  \label{equ:nabla_P_I_sigma}
    \nabla_{\theta}L = \nabla_{\theta} J - \sigma\mathbb{E}_{\tau\in\mathcal{B}^{i}}\left[\operatorname{d}(\tau, \mathcal{M}^{i})\nabla_{\theta}\log p_{\theta}(\tau)\right].
\end{equation}
where $p_{\theta}$ is a distribution induced by $\pi_{\theta}$ over the trajectory space, and $p_{\theta}$ represents the probability of the trajectory $\tau$. $p_{\theta}$ can be expressed in terms of the environment dynamics model and the policy of the agent, i.e. $p_{\theta}(\tau)=\rho(s_0)\sum_{t=0}^{T}\pi_{\theta}(a_t\vert s_t)p(s_{t+1}\vert s_t,a_t).$ Therefore, the gradient of the score function of the trajectory distribution has the expression $\nabla_{\theta}\log p_{\theta}(\tau)=\sum^{T}_{t=0}\log\pi_{\theta}(a_t\vert s_t)$, and it does not contain the environment dynamic model.

\subsection{Policy Exploration}
\label{sec:imporoving exploration}
As described in Sec.~\ref{sec: soft self-imitation learning}, if an agent has collected specific trajectories when interacting with the environment and stores them in the trajectory buffer, our method will regard these trajectories as guidance in the policy improvement step and direct the agent to revisit the regions where past good trajectories are located gradually. However, it might cause the agent to get stuck in local optima. To achieve better exploration performance in challenging tasks with sparse and deceptive rewards, we employ a group of heterogeneous agents to interact with the environment simultaneously. We hope to enable different agents on the team to reach diverse regions of the state space and drive deep exploration.

To maintain the diversity of the team, we first introduce a novel measurement of diversity. Considering the continuous control tasks we focus on, we use the mean MMD distance between the agents as the diversity measurement. The mean MMD distance is computed with current trajectories collected by agents and their mean trajectories, where the mean trajectories are generated by performing the means of Gaussian action distributions produced by agents. Specifically, let $\bar{\tau}^{i}$ denote the mean trajectories of agent $i$, let $\Pi$ denote the ensemble of agents we employ, and let $\mathcal{S}$ be the set of mean trajectories of all agents. The diversity measurement of an agent is calculated as follows:
\begin{equation} 
  \label{equ:diversity_conti}
  \begin{aligned}
    \operatorname{div}(\Pi)=\frac{1}{\vert\Pi\vert}\sum_{i=1}^{\vert\Pi\vert}D_{\mathrm{MMD}}(\pi^{i},\Pi\backslash\pi^{i}),
  \end{aligned}
\end{equation} 
where $D_{\mathrm{MMD}}(\pi^{i},\Pi\backslash\pi^{i})$ is defined as:
\begin{equation}
 D_{\mathrm{MMD}}(\pi^{i},\Pi\backslash\pi^i)=\min_{\bar{\tau}^{j}\in\mathcal{S}\backslash\bar{\tau}^{i}}\mathbb{E}_{\tau\in\mathcal{B}^{i}}\left[\mathrm{MMD}(b(\tau), b(\bar{\tau}^j))\right].
\end{equation}

For the discrete control tasks, we utilize the current optimal trajectories to compute the diversity measurement of agent team $T$, and the optimal trajectories are produced by agents through outputting actions with the highest $Q$-values every time step. Finally, the objective function of policy exploration is given as:
\begin{equation} 
  \label{equ:policy_explore}
  \begin{aligned}
    \theta_{k+1}=&\max_{\theta}\, \operatorname{div}(\Pi_{\theta_{k+\frac{1}{2}}}), \\
    \mathrm{s.t.}\quad D_{\mathrm{KL}}(\pi^{i}_{k+1}&,\pi^{i}_{k+\frac{1}{2}})\le\delta, i\in{1,\dots,\vert\Pi\vert}.
  \end{aligned}
\end{equation} 

Intuitively, the policy exploration step prevents the team of agents from being stuck in the same local optimum by driving different agents to reach diverse regions of the state space. This is achieved by finding an ensemble of policies that maximizes the diversity measurement defined in Eq.~\eqref{equ:diversity_conti}. In the meantime, we require that every new policy $\pi_{k+1}$ after the policy update lies inside the trust region around the old policy $\pi_{k_+\frac{1}{2}}$, which is defined as $\{\pi: D_{\mathrm{KL}}(\pi,\pi_{k+\frac{1}{2}})\le\delta\}$, which can avoid suffering severe performance degradation.

\textbf{Optimization Process for the Diversity Objective}\quad 
To solve the constrained optimization problem, we propose to solve it by linearizing around current policies $\pi^{i}_{k},i=1,\dots,\vert\Pi\vert$. The gradient of diversity measurement in Eq.~\eqref{equ:policy_explore} with respect to the policy parameters can be easily calculated with the mean trajectories and agent rollouts. Denoting the gradient of diversity measurement $g$, the Hessian matrix of the KL-divergence for agent $i$ as $H^i$, the linear approximation to Eq.~\eqref{equ:policy_explore} is as follows:
\begin{equation} 
  \label{equ:policy_explore_appro}
  \begin{aligned}
    \theta_{k+1}=&\max_{\theta}\, g^{T}(\theta-\theta_{k}), \\
    \mathrm{s.t.}\ \frac{1}{2}(\theta-\theta_k)^T H^i_{k}&(\theta-\theta_k)\le\delta, \ i\in{1,\dots,\vert\Pi\vert}.
  \end{aligned}
\end{equation} 

Then, we adopt the conjugate gradient method~\cite{Schulman2015TrustRP} to approximately compute the inverse of $H_{k}^i$ and the gradient direction. However, the trust-region optimization can incur slow update of parameters and thus reduce the sample efficiency and improve the computational cost. In the beginning phase of the training process, we can use a first-order optimization method like~\cite{schulman2017proximal} to solve the optimization problem in Eq.~\eqref{equ:policy_explore}.

\section{Evaluation of Results}
\label{sec:experience}
This section presents our experimental results and compares our method's performance with other baseline methods. In section~\ref{sec:setup}, we present an overview of experiment setups we used to evaluate our methods. Section~\ref{subsec:performance_gridworld} \textasciitilde~\ref{subsec:performance_Mujoco} report the results in different experiment environments, respectively.

\subsection{Experimental Setup}
\label{sec:setup}

\subsubsection{Environments}
\label{subsubsec:environments}
\begin{figure}[htb]
  \centering
  \subfigure[]{
    \includegraphics[width=4.cm]{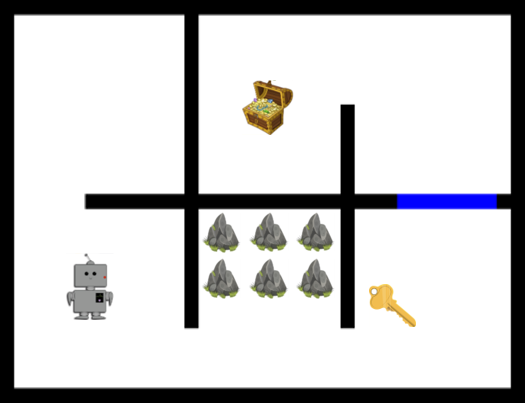}
    \label{fig:sparse_maze}
  }
  \subfigure[]{
    \includegraphics[width=4.cm]{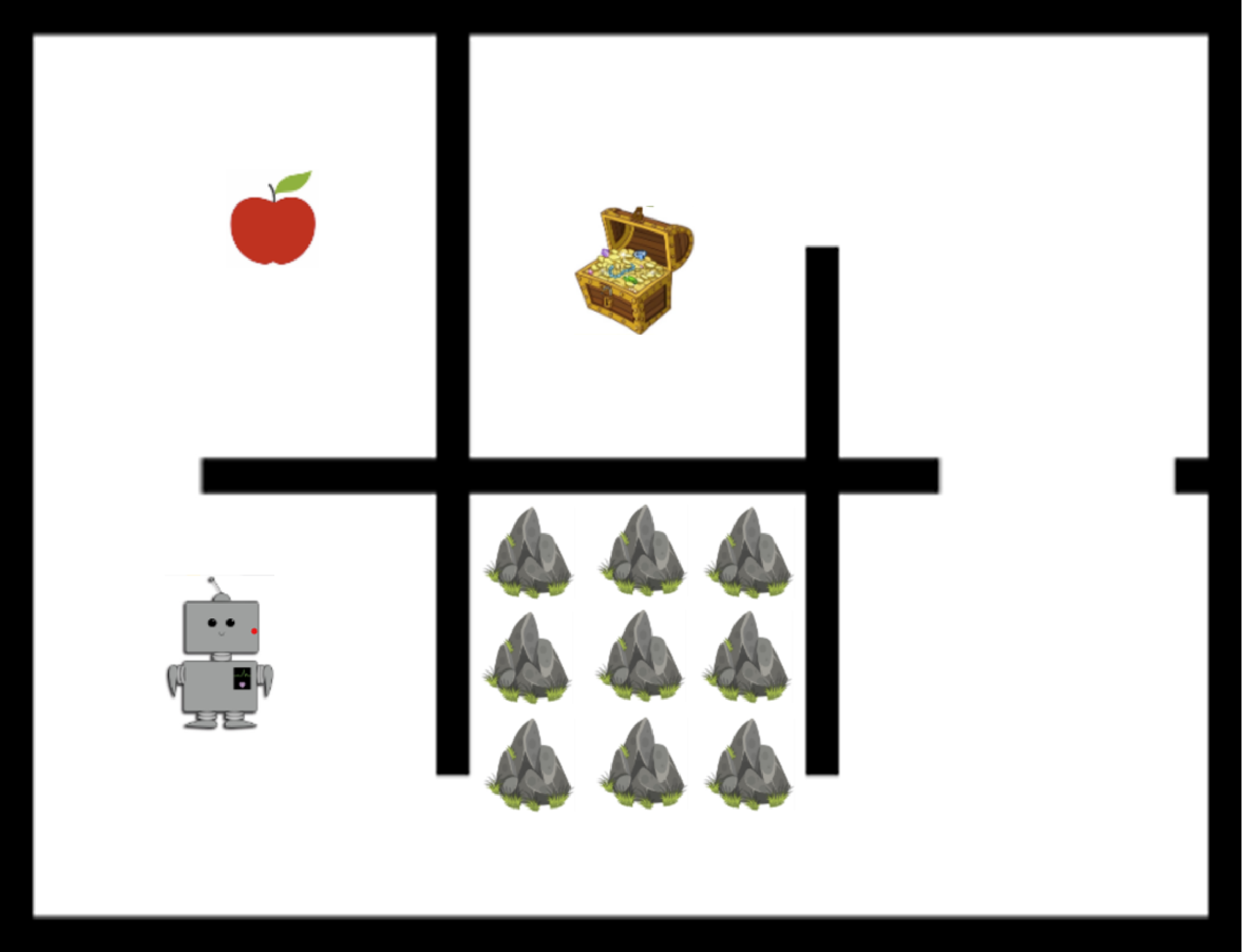}
    \label{fig:deceptive_maze}
  }
  \caption{A collection of environments with discrete state-action spaces that we use.
    (a) Huge grid-world maze with sparse rewards: Key-Door-Treasure domain.
    The agent should pick up the key (+2) in the right-down room in order to open the blue door (+4) and 
    collect the treasure (+4) in the middle-up room to maximize the reward.
    (b) Huge grid-world maze with deceptive rewards.
    There is an apple in the left-up room that gives small rewards (+2) and a treasure in the middle-up 
    room which generates the higher rewards (+10).}
  \label{fig:mazes}
\end{figure}

\textbf{Grid world.} To illustrate our method's effectiveness, we design a huge 2D grid-world maze based on Gym~\cite{1606.01540} with two different settings: sparse rewards and deceptive rewards. At each time step, the agent observes its coordinates relative to the starting point and chooses from four possible actions: move east, move south, move west, and move north. At the start of each episode, the agent starts from the bottom-left corner of the map, and an episode terminates immediately once a reward is collected. In the sparse reward setting shown in Fig~\ref{fig:mazes}\subref{fig:sparse_maze}, there is only a single goal located at the top-right corner, and the agent will be rewarded with 6 when reaching this goal. On the other hand, in the deceptive reward setting depicted in Fig~\ref{fig:mazes}\subref{fig:deceptive_maze}, there is a misleading goal with a reward of 2 in the upper left room of the grid world that can lead the agent to local optima.

\begin{figure}[htb]
  \centering
  \begin{minipage}{.40\textwidth}
    \centering
    \subfigure[]{\label{fig:swimmer_maze}
      \includegraphics[width=4.5cm]{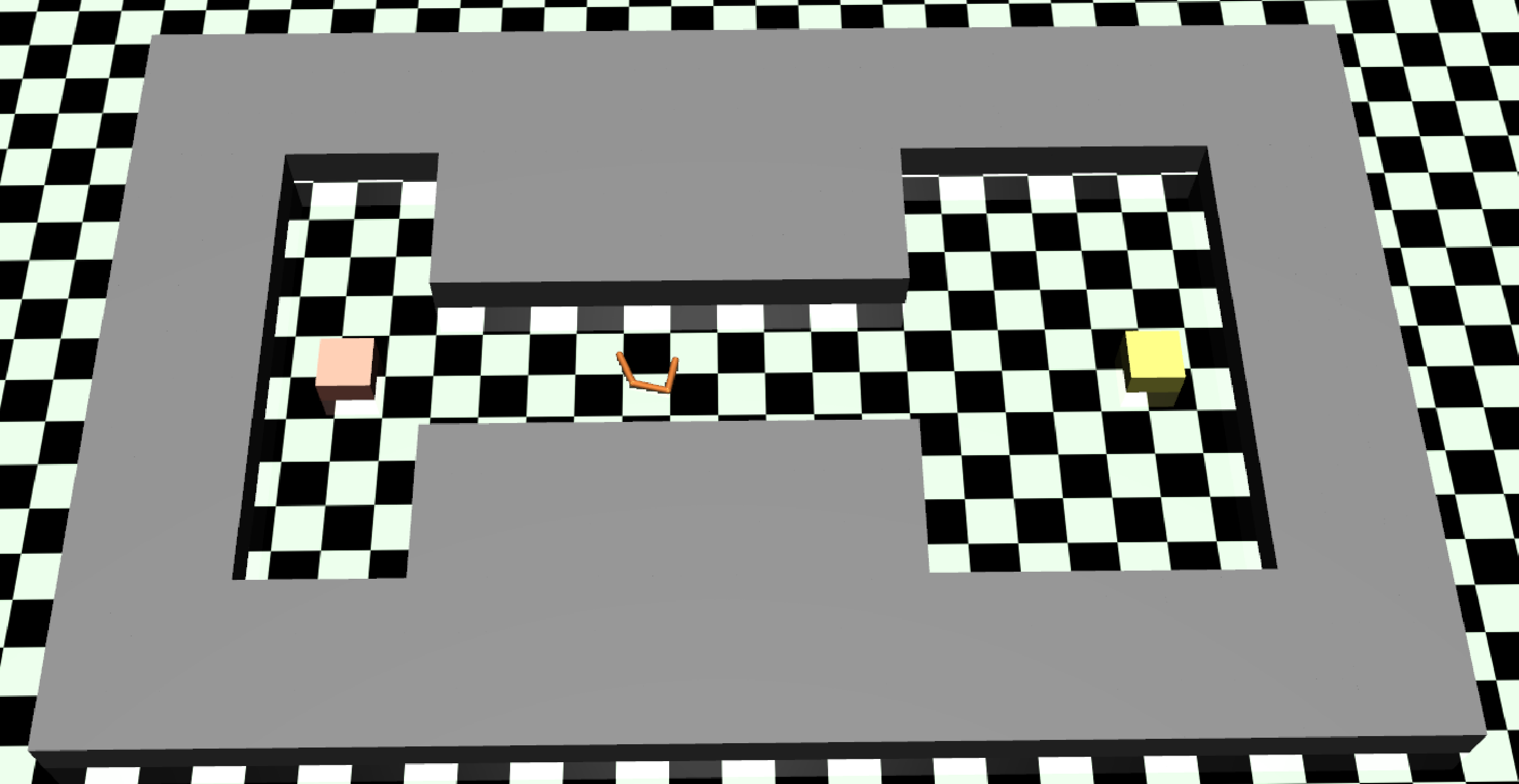}
    }
  \end{minipage}
  \begin{minipage}{.40\textwidth}
    \flushright
    \subfigure[]{\label{fig:Ant_maze}
      \includegraphics[width=4.cm]{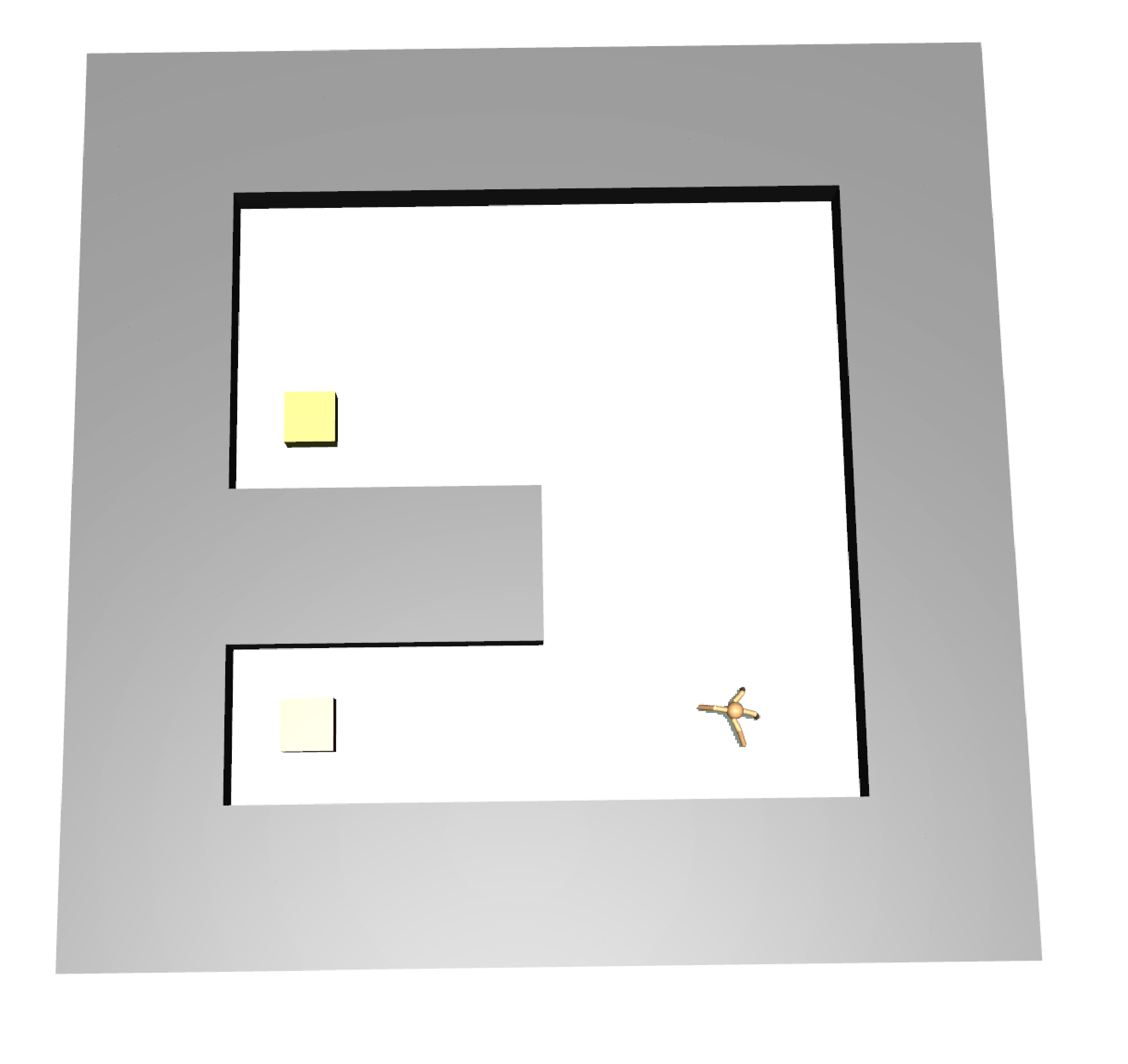}
    }
  \end{minipage}
  \caption{A collection of environments with continuous state-action spaces that we use: (a) Swimmer in the maze, (b) Ant in the maze.}
  \label{fig:mujoco_mazes}
\end{figure}

\textbf{MuJoCo.} For continuous control tasks, we also conduct different experiments in environments based on MuJoCo physical engine~\cite{todorov2012mujoco}. Two continuous robotic control tasks, ant and swimmer, are selected to evaluate our methodology's and baseline methods' performance. In each task, the agent takes a vector of the physical state containing the agent's joint angles and task-specific attributes as the input of the policy. Examples of task-specific attributes include goals walls and sensor readings. Then, the control policy generates a vector of action values that the agent performs in the environment. We compare our method to previous algorithms in the following tasks:
\begin{itemize}
  \item Swimmer Maze: The swimmer is rewarded for reaching the goal positions in the maze shown in Fig.~\ref{fig:mujoco_mazes}\subref{fig:swimmer_maze}. The agent can obtain sub-optimal rewards (+200) by arriving at the left goal, while the agent is rewarded with 500 as it reaches the goal on the right of the maze. 
  \item Ant Maze: The ant is rewarded for arriving at the specified positions in the maze shown in Fig.~\ref{fig:mujoco_mazes}\subref{fig:Ant_maze}. The ant can collect small rewards when reaching the goal below the maze and maximize the reward if it reaches the goal up the maze. 
\end{itemize}

\subsubsection{Baseline Methods}
\label{subsubsec:baselines}
The baseline methods used for performance comparison vary in different tasks. For discrete and continuous control tasks, we compare our algorithm with the following baseline methods: 
(1) PPO~\cite{schulman2017proximal}, (2) SAC~\cite{haarnoja2018soft},
(3) DPPO: Distributed PPO~\cite{heess2017emergence},
(4) Div+A2C: A2C with a distance measure regularization~\cite{hong2018diversity},
(5) PPO+SIL: PPO with Self-Imitation Learning~\cite{oh2018self},
(6) PPO+EXP: PPO with count-based exploration bonus augmented reward function: $r(s, a)+\frac{\lambda}{\sqrt{N_e}}$, where $N_e$ is the number of times the state cluster under the state representation $e$ is visited during training and $\lambda$ is the hyperparameter that controls the weight of the exploration bonus term. For each baseline, we adopt the parameters that produce the best performance during the parameter search, and not all baselines are adopted in each task.

\subsection{Performance in the Huge Grid-World Maze}
\label{subsec:performance_gridworld}
In this experiment, we evaluate methods in 2D mazes with different reward settings: sparse and deceptive rewards. We consider five baseline methods in these experiments: A2C, PPO, PPO+SIL, PPO+EXP, and Div-A2C. The performances of each method are reported in terms of average return and success rate learning curves. All learning curves are averaged over 8 runs.
\begin{figure}[htb]
  \centering
  \subfigure[]{
    \includegraphics[width=5.7cm]{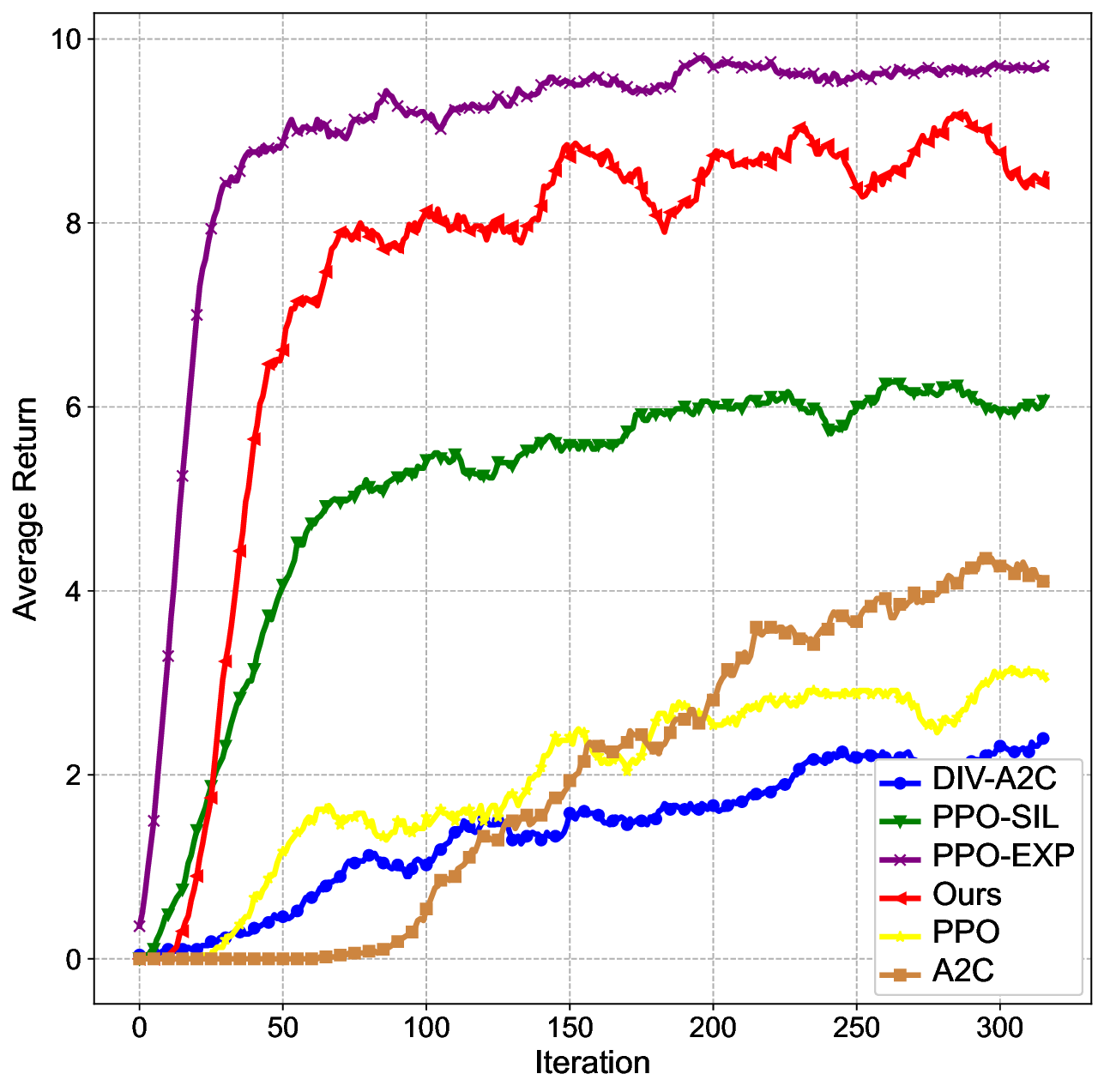}
    \label{fig:discrete_sparse_average}
  }
  \subfigure[]{
    \includegraphics[width=5.7cm]{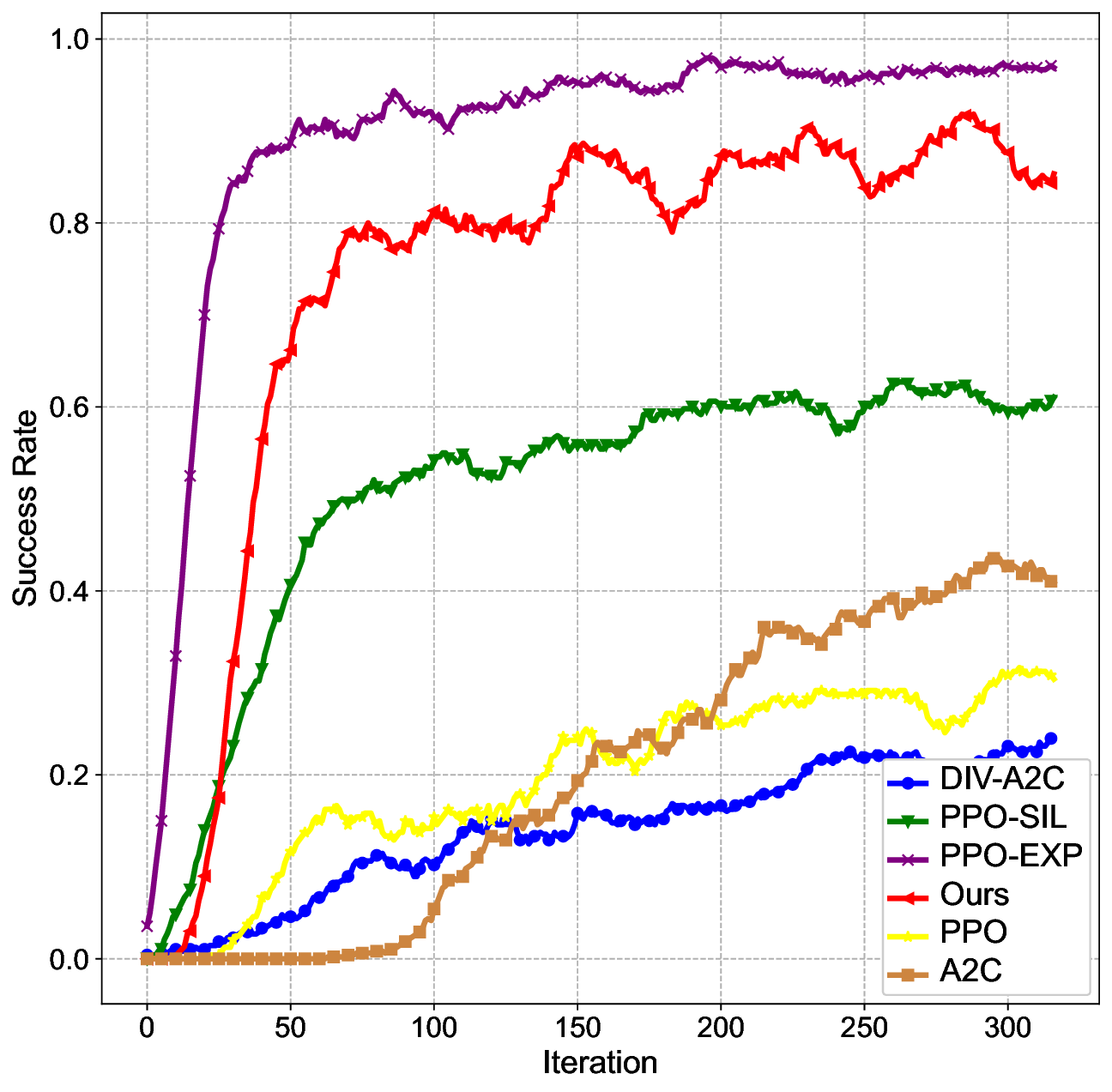}
    \label{fig:discrete_sparse_success}
  }
  \caption{Learning curves of average return and success rate in the huge grid-world maze with sparse rewards. Specifically, the success rate is used to illustrate the frequency at which agents reach the globally optimal goal during the training process.}
  \label{fig:discrete_sparse_performance}
\end{figure}

\textbf{Sparse reward.} Fig.~\ref{fig:mazes}~\subref{fig:sparse_maze} shows the structure of the sparse reward grid-world maze. We implement our method based on the PPO~\cite{schulman2017proximal} algorithm in this maze environment with the discrete state-action space. In this experiment we select A2C, vanilla PPO, PPO+SIL, Div-A2C, as well as PPO+EXP as the baselines. We adopt the settings that produce the highest performance for each method during the hyperparameter search. The learning curves are presented in Figure~\ref{fig:discrete_sparse_performance}. Compared with other baseline methods, we notice that our method is able to learn the optimal policy at a high learning rate and achieve a good performance evaluation of average return and success rate in this task. Our method can encourage the agent to visit the state space's underexplored regions and better use past good trajectories maintained in the memory buffers. Therefore, our method can prompt the agent to focus on the state space from which the agent can collect the sparse rewards with a higher probability. The PPO and A2C agents do not have any specially designed mechanism for exploration; hence, they only arrive at the treasure and obtain the sparse rewards with a very low probability. These agents are not able to learn the optimal policy leading to the treasure due to the low percentage of valid samples. In the training process, the PPO+EXP agent can explore the environment better and occasionally collect the treasure to achieve the best episode return. We also note that it is difficult for the Div-A2C method to encounter the sparse rewards in one short burst. While the PPO+SIL agent can utilize the past good trajectories, this method's success rates and average returns are lower than ours at the end of the training process.
\begin{figure}[htb]
  \centering
  \subfigure[]{
    \includegraphics[width=5.6cm]{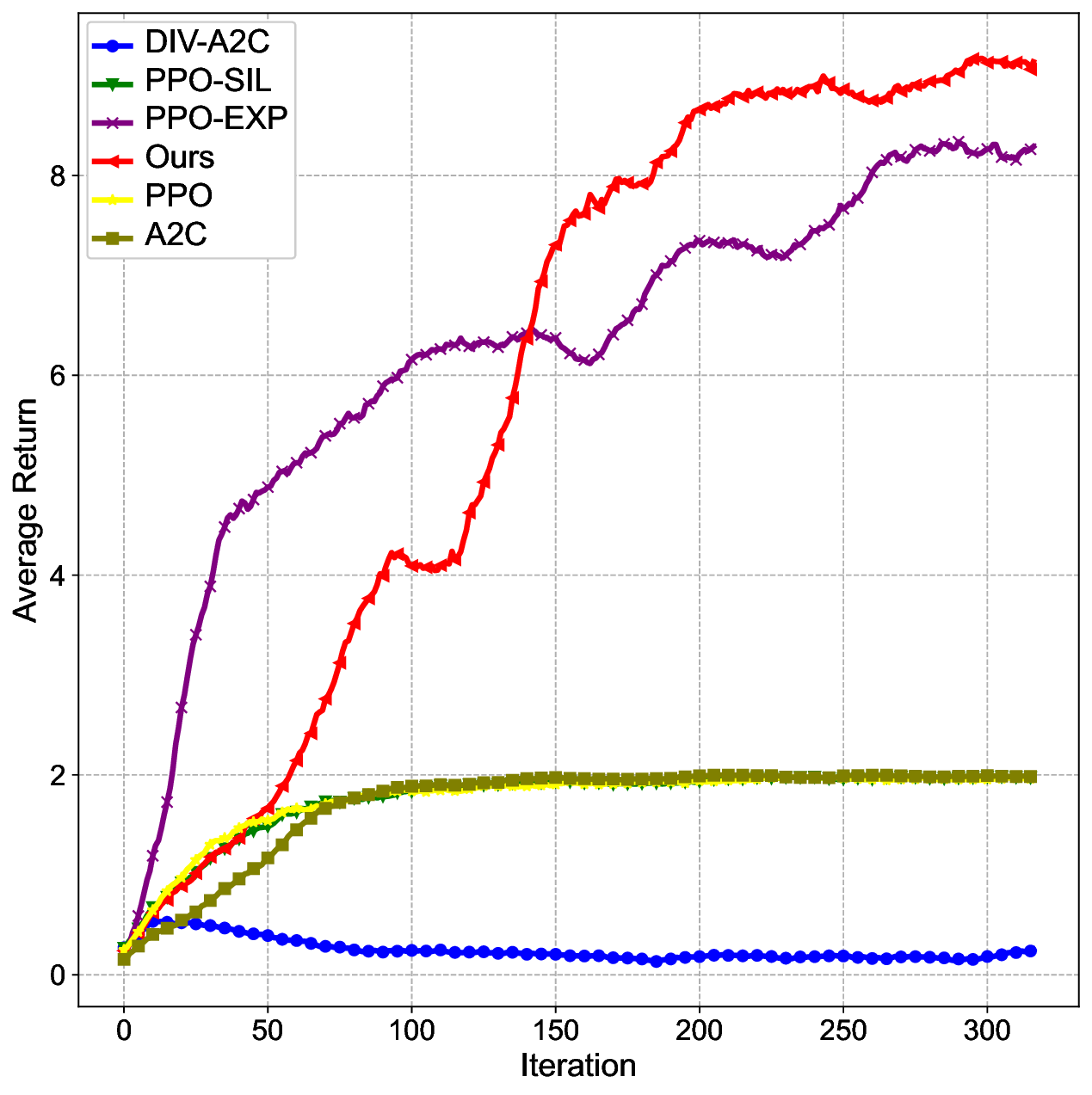}
    \label{fig:discrete_deceptive_average}
  }
  \subfigure[]{
    \includegraphics[width=5.7cm]{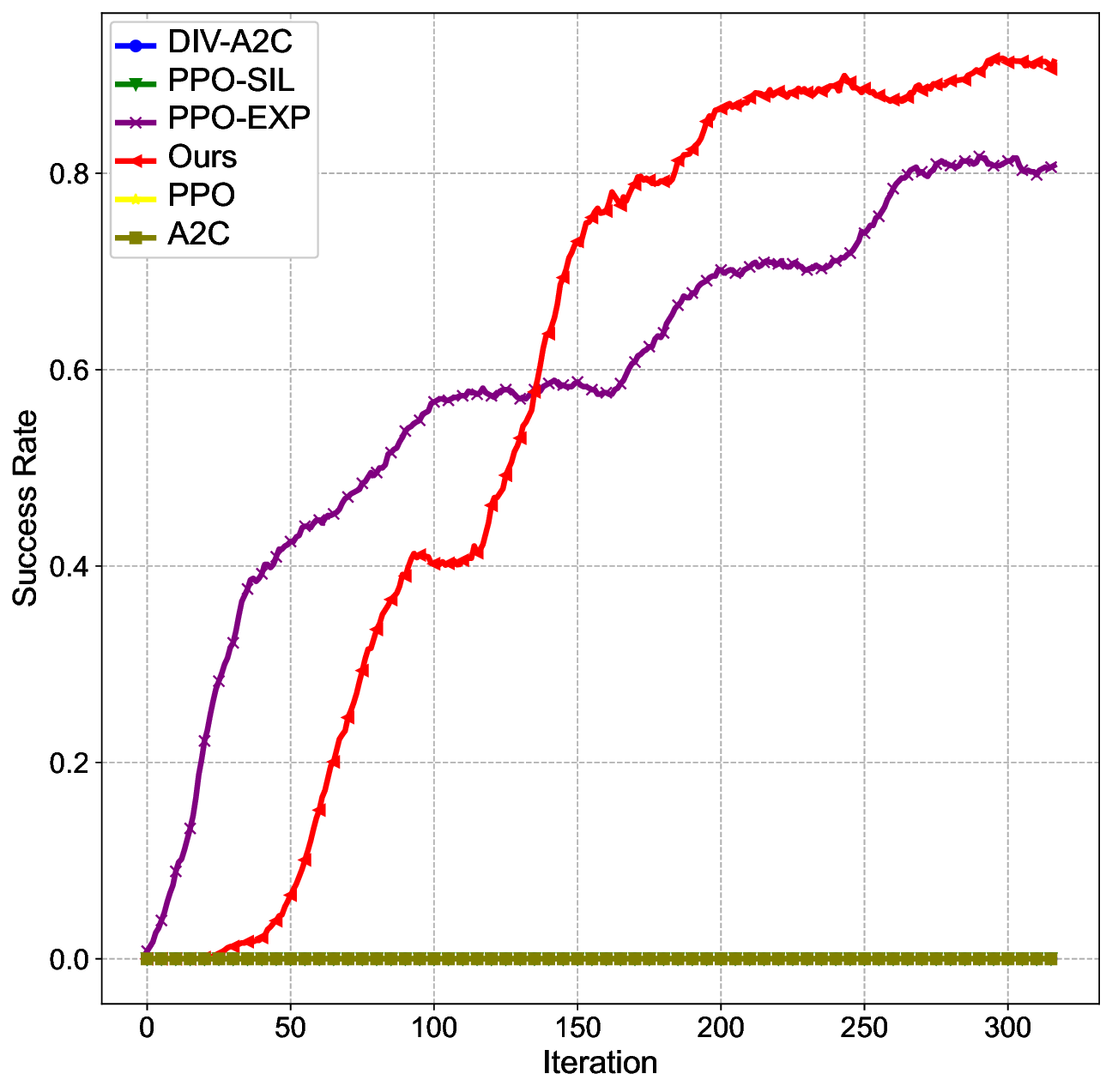}
    \label{fig:discrete_deceptive_success}
  }
  \caption{Learning curves of average return and success rate in the huge grid-world maze with deceptive rewards. Specifically, the success rate is used to illustrate the frequency at which agents reach the globally optimal goal during the training process.}
  \label{fig:discrete_deceptive_performance}
\end{figure}

\textbf{Deceptive reward.}
Fig.~\ref{fig:mazes}\subref{fig:deceptive_maze} illustrates the deceptive grid-world maze. From Fig.~\ref{fig:discrete_deceptive_performance}, we notice that the average rewards of the baseline methods have a slight increase compared with those in the sparse reward settings. These methods, except for PPO+EXP, can only achieve the suboptimal rewards by collecting the apple, and hence the agents are stuck with the suboptimal policy due to deceptive rewards. In contrast, our method not only adopts myopic and suboptimal behaviors but also treats the past good trajectories as guidance and allows the agent to reach diverse regions in the state space by improving upon past trajectories to generate good new trajectories. Although PPO+EXP can reach the treasure and obtain the highest rewards, its learning process has greater instability, leading to inferior performance.

\begin{figure}[htb]
  \centering
  \subfigure[PPO]{
    \includegraphics[width=3.5cm]{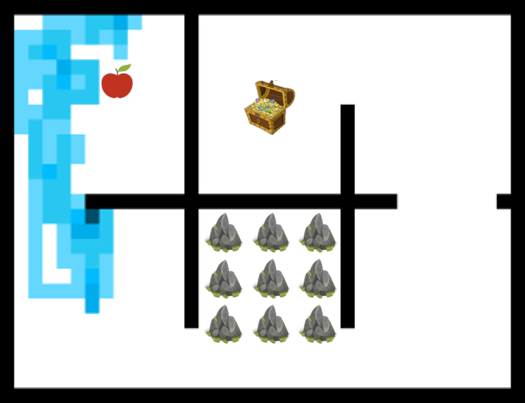}
    \label{fig:ppo}
  }
  \subfigure[A2C]{
    \includegraphics[width=3.5cm]{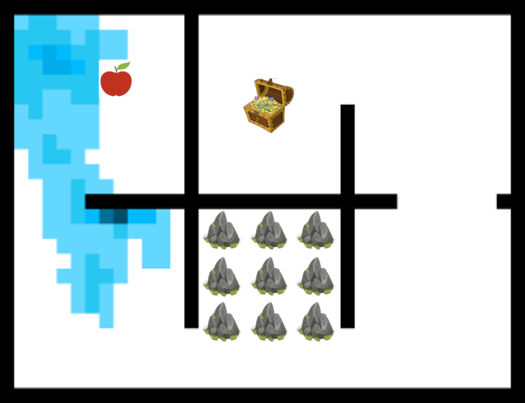}
    \label{fig:a2c}
  }
  \subfigure[Div-A2C]{
    \includegraphics[width=3.5cm]{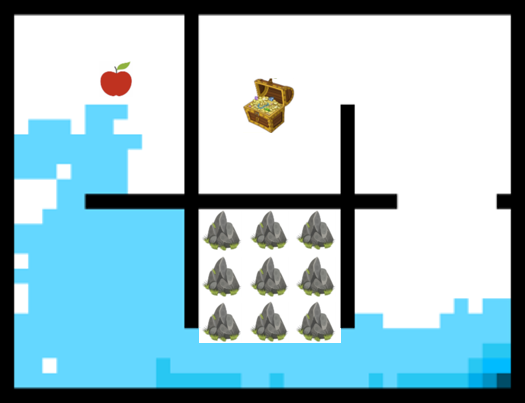}
    \label{fig:a2c_div}
  }

  \subfigure[PPO-SIL]{
    \includegraphics[width=3.5cm]{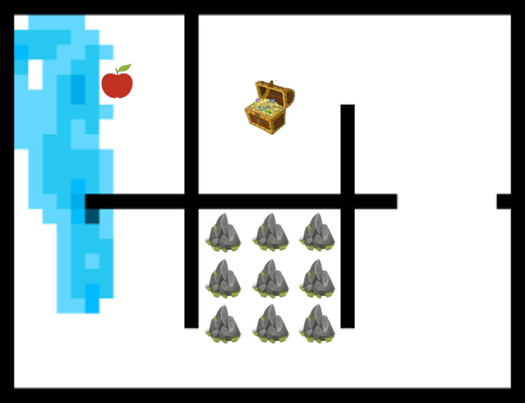}
    \label{fig:ppo_sil}
  }
  \subfigure[PPO-EXP]{
    \includegraphics[width=3.5cm]{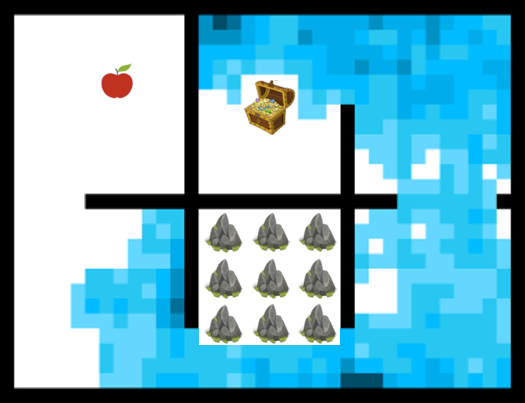}
    \label{fig:ppo_exp}
  }
  \subfigure[Ours]{
    \includegraphics[width=3.5cm]{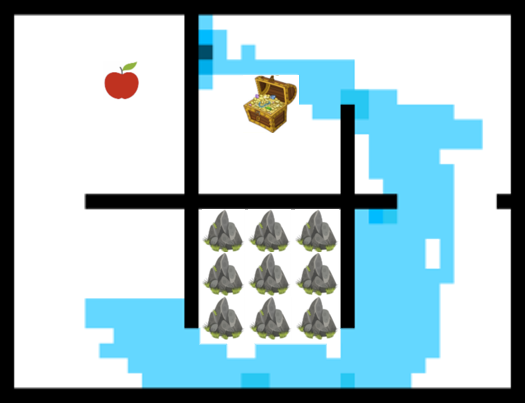}
    \label{fig:ours}
  }
  \caption{State-visitation counts for different algorithms: (a) PPO, (b) A2C, (c) Div-A2C, (d) PPO-SIL, (e) PPO-EXP, (f) Ours. PPO, A2C and PPO-SIL are easily trapped into local optimum. Div-A2C can visit regions where the apple is not located but can not arrive at the treasure and obtain the optimal reward. PPO+EXP enables the agent to reach the treasure and obtain the highest rewards; however, it spends amounts of computation to visit parts of state-action space where it cannot obtain any useful rewards. Our algorithm can explore the state-action space systematically and collect the optimal rewards quickly, which enables the agent to learn the optimal policy at a high learning rate}
  \label{fig:state_visitation}
\end{figure}

As shown in Fig.~\ref{fig:state_visitation}, we plot the state-visitation counts for all methods in the maze with deceptive rewards, which explicitly illustrate how different agents explore the 2D grid world environments. From the state-visitation count graph (see Fig.~\ref{fig:state_visitation}), it can be seen that the four baseline approaches are either prone to falling into a local optimum or they cannot explore the environment sufficiently to visit the goal with a larger reward. PPO+EXP can obtain optimal rewards from the treasure, but it requires considerable computation to visit the meaningless region of the state action space due to intrinsic rewards. However, it can be observed from Fig.~\ref{fig:state_visitation}\subref{fig:ours} that our POSE method is able to escape from the area of deceptive rewards and explore a wider and farther region of the 2D grid world significantly, and arrive at the goal with the optimal reward successfully, without spending much time visiting the insignificant region of the state space.

\subsection{Performance Comparison in MuJoCo Environments}
\label{subsec:performance_Mujoco}
\begin{figure}[htb]
  \centering
  \subfigure[]{
    \includegraphics[width=5.7cm]{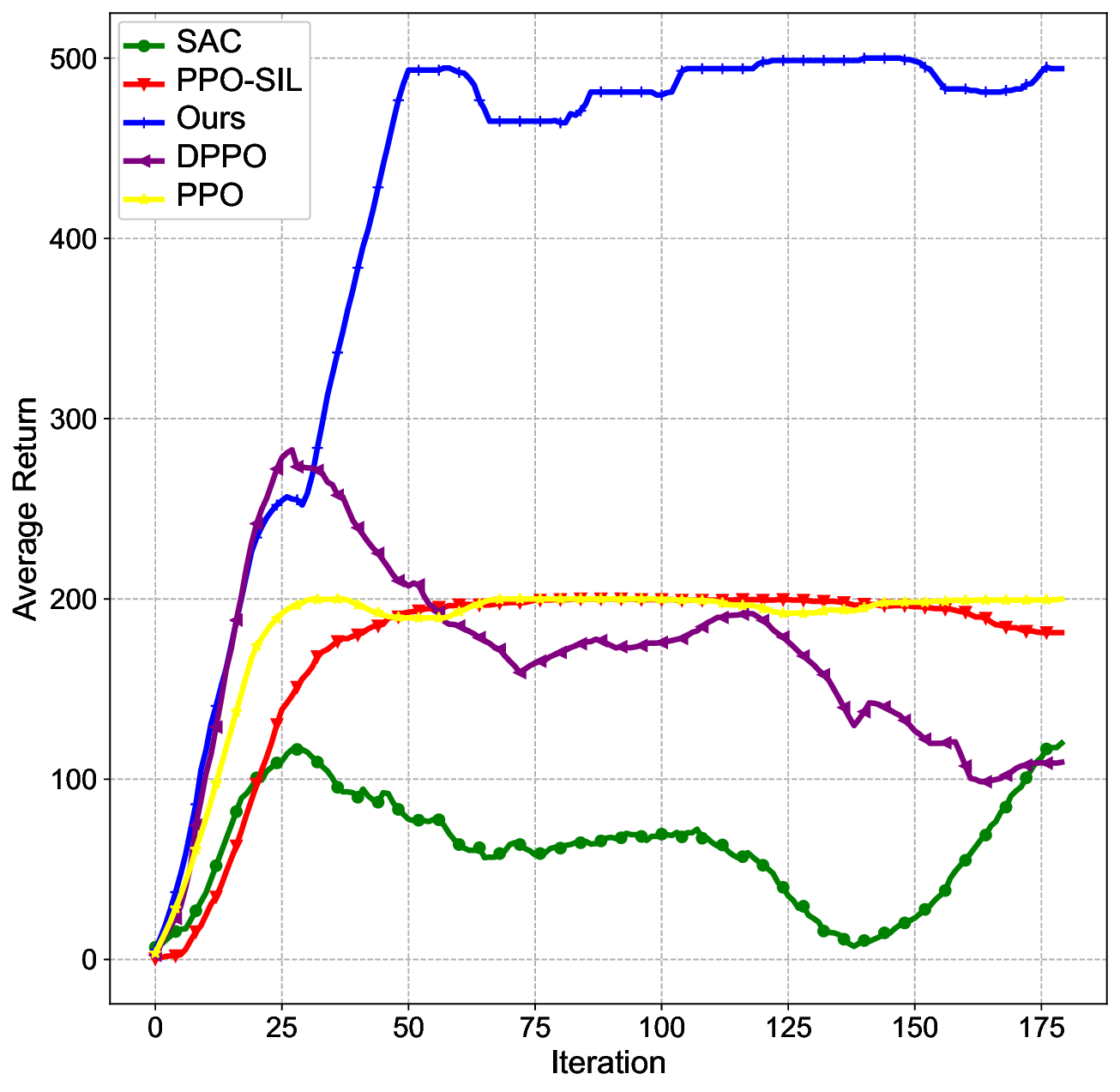}
    \label{fig:swimmer_average}
  }
  \subfigure[]{
    \includegraphics[width=5.7cm]{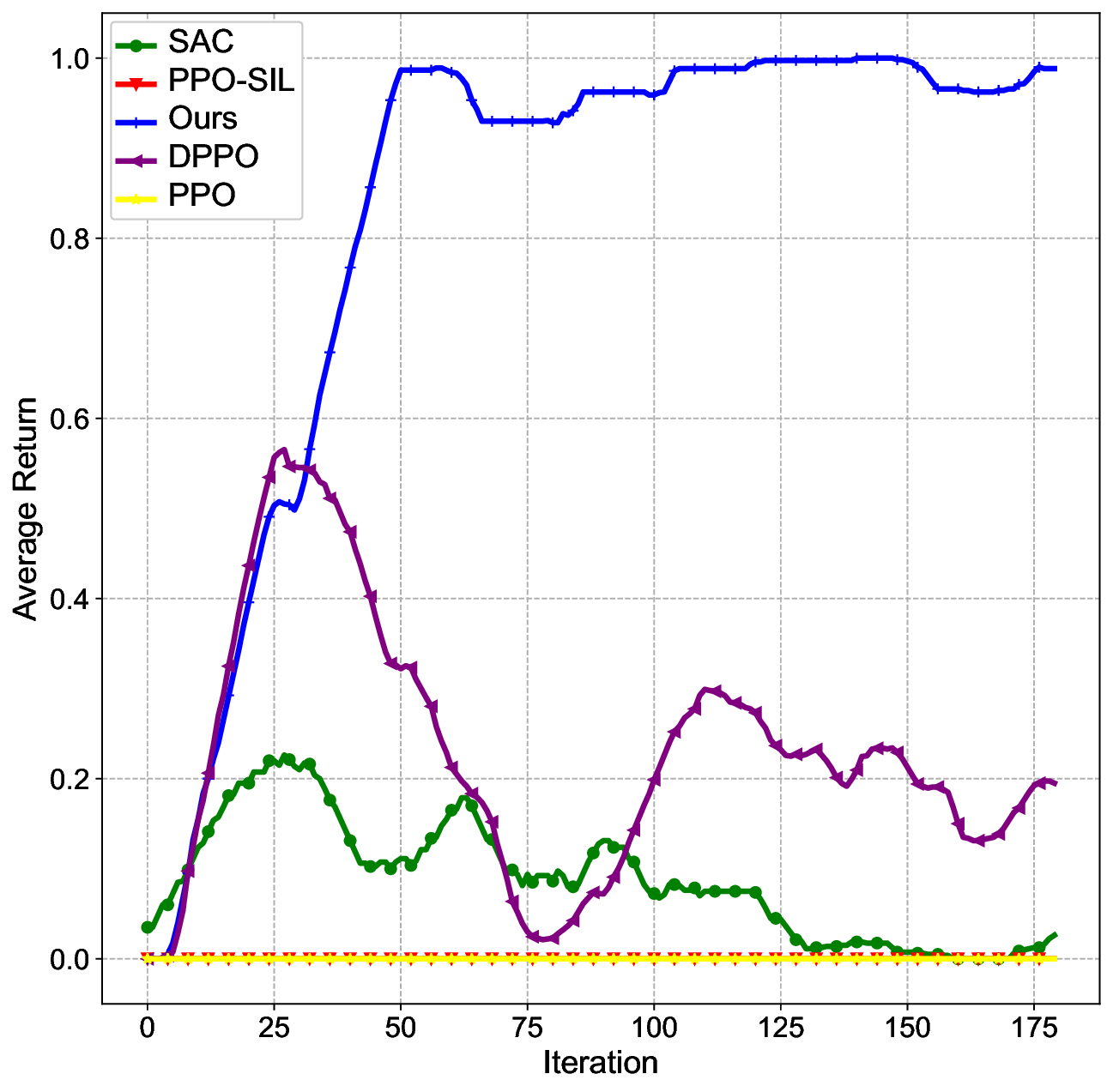}
    \label{fig:swimmer_success}
  }
  \caption{Learning curves of average return and success rate in Swimmer Maze. Specifically, the success rate is used to illustrate the frequency at which agents reach the globally optimal goal during the training process.}
  \label{fig:swimmer_results}
\end{figure}

\begin{figure}[htb]
  \centering
  \subfigure[]{
    \label{fig:ant_average}
    \includegraphics[width=5.7cm]{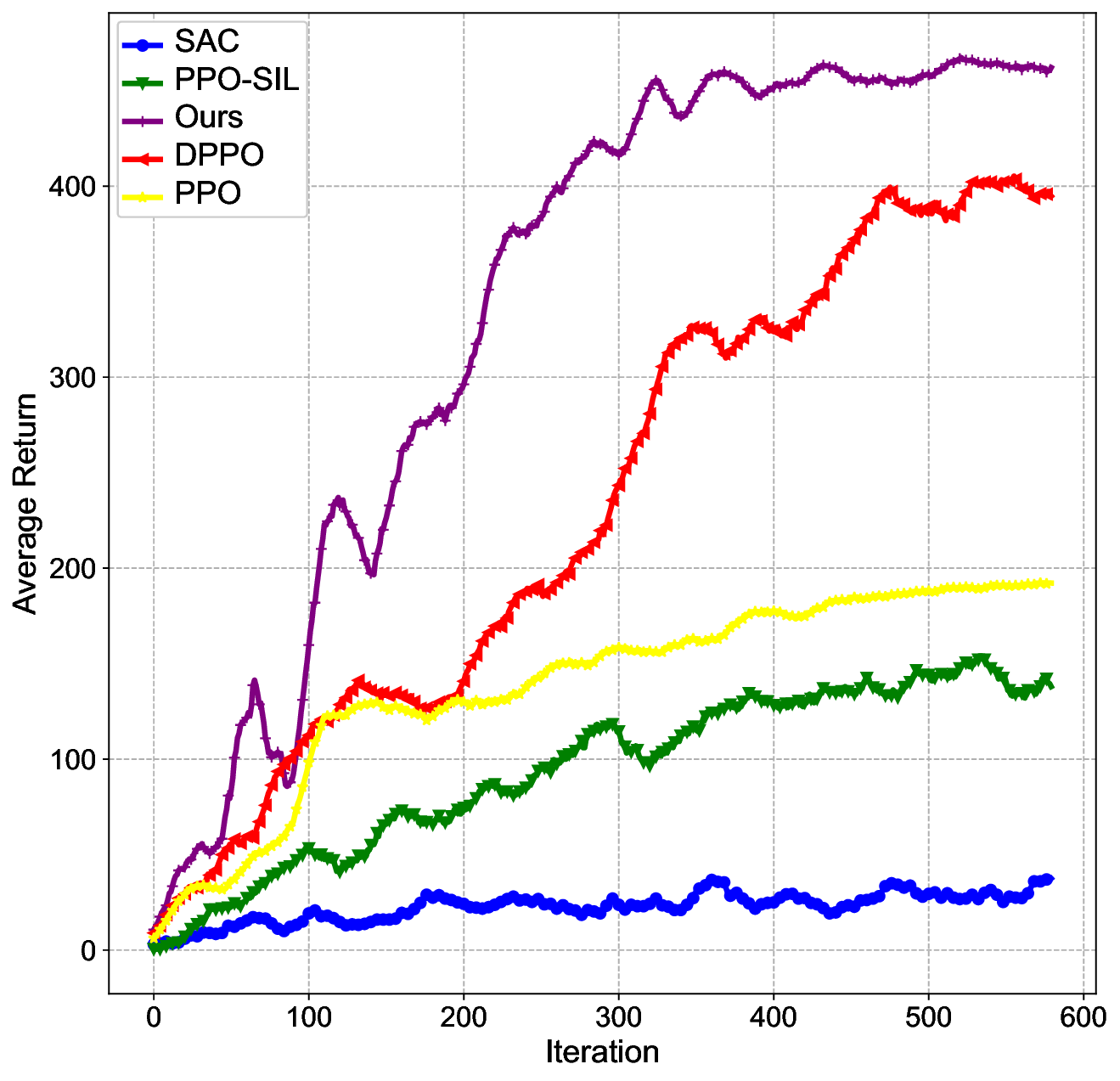}
  }
  \subfigure[]{
    \label{fig:ant_success}
    \includegraphics[width=5.7cm]{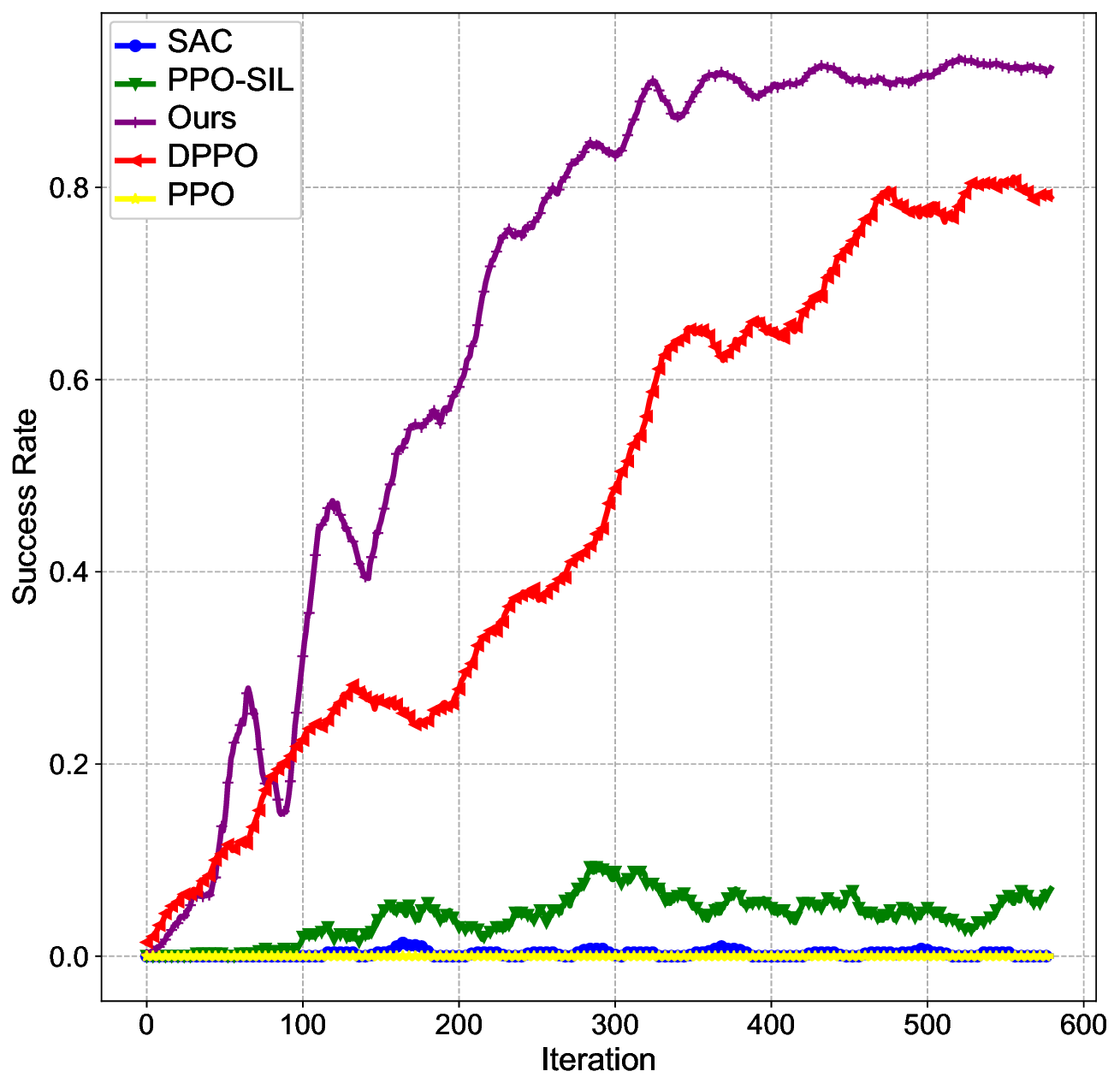}
  }
  \caption{
    Learning curves of average return and success rate in Ant Maze. Specifically, the success rate is used to illustrate the frequency at which agents reach the globally optimal goal during the training process.
  }
  \label{fig:ant_result}
\end{figure}

We evaluate our method on continuous control tasks shown in Fig.~\ref{fig:mujoco_mazes} with sparse and deceptive rewards based on MuJoCo physical engine and similarly plot the in-training median scores in Fig.~\ref{fig:swimmer_results} and~\ref{fig:ant_result}. We consider four baseline methods in these experiments: SAC, PPO, PPO+SIL, and DPPO. The performance of each method is reported in terms of average return and success rate learning curves. Like the discrete maze experiments, all learning curves are averaged over five runs with different random seeds.

Compared with baseline methods, it is observed that our method can learn considerably faster and obtain higher average returns and success rates. The average returns and success rates of POSE and other baseline methods in the Swimmer Maze are usually higher than those in the Ant Maze because the dimensions of the state and action space of the swimmer are lower than the ant, and the swimmer will not trip over due to incorrect action inputs. POSE reaches a success rate of almost 100\% in less than 200 epochs.

While PPO and PPO+SIL often adopt myopic behaviors and converge to suboptimal policies, POSE is able to eliminate the local optimum and find better strategies for the larger episode returns. We also compared our algorithm with the state-of-the-art RL method SAC and DPPO. SAC is based on the maximum entropy RL framework, which trains a policy by maximizing the trade-off between expected return and entropy. As a result, SAC cannot achieve a significantly better performance of exploration in our experiments than PPO. It rarely encounters the optimal rewards received from the treasure and occasionally gathers trajectories leading to the treasure in the swimmer maze. Thus, this off-policy method might forget the past good experience and fail to learn the optimal policy to achieve the best reward. DPPO can learn the policy leading to the optimal reward, but this method has a lower learning rate and success rate. In contrast, our POSE method can also successfully generate trajectories visiting novel states and save the trajectories with high returns in the buffer. The POSE agent is able to replicate the past good experience by using the highly rewarded trajectories as guidance and learn the optimal policies with the highest learning rate and success rate.

\section{Conclusion}
\label{sec:conclusion}
In this paper, we study the problem of how to design the practical RL algorithm in tasks where only sparse and deceptive feedback is provided. We propose a novel two-step policy optimization framework called POSE, which exploits diverse and imperfect demonstrations for faster and more efficient online RL. By considering diverse past trajectories as soft guidance, the agent can reproduce the trajectories easily and smoothly move beyond to find near-optimal policies, which regulates the direction of policy improvement and accelerates the learning speed. Furthermore, a novel diversity measure is introduced to direct each agent on the team to visit the different underexplored regions of the state space and drive deep exploration. Experimental results on physical control benchmarks demonstrate the effectiveness of our approach over other baseline methods for the efficient exploration and avoidance of local minima in tasks with long horizons and sparse or deceptive rewards.

\backmatter



\section*{Declarations}

\begin{itemize}
\item \textbf{Funding}
This work was supported by the National Key R\&D Program of China (2022ZD0116401)
\item \textbf{Conflict of interest} 
The authors declare that there is no conflict of interest.
\item \textbf{Availability of data and materials}
The datasets generated during and/or analyzed during the current study are available from the corresponding author upon reasonable request.
\end{itemize}






\begin{appendices}
  \section{Algorithm Training Process}
  \label{appendix:A}
  Algorithm~\ref{algo:DS2L} describes our method in detail. At each iteration, the algorithm is executed according to the framework shown in Fig~\ref{fig:diverse_exploration}, and experiences generated by each agent in the environment are stored in their respective on-policy training batches. Then we use the on-policy trajectory data to update the soft self-imitation learning batches $\mathcal{M}$ according to Section~\ref{sec:overview}. Furthermore, we compute the advantage of the current policy and the distance between the current trajectory $\tau$ and the soft self-imitation replay buffer $\mathcal{D}_i$ for each agent, and we utilize them to estimate the gradient of $\nabla_{\theta} \tilde{L}$ in Equation~\eqref{equ:nabla_P_I_sigma}. Finally, we update the parameters of $ \pi_{\theta} $ with the gradient ascent algorithm and adapt the penalty factor according to the MMD distance.
  \renewcommand{\algorithmicrequire}{\textbf{Input:}}
  \renewcommand{\algorithmicensure}{\textbf{Output:}}
  \begin{algorithm}[htb]
    \caption{POSE}
    \label{algo:DS2L}
    \begin{algorithmic}[1]
      \Require number of agents $N$, learning rate $\alpha$, on-policy training buffer $\mathcal{R}_i$ for each agent, highly-rewarded trajectory buffer $\mathcal{D}_i$ for each agent, sequence length $m$, and number of epochs $M$.
      \State Initialize policy weights $\theta$ of each agent.
      \State Initialize the prior good trajectory buffer $\mathcal{M}_\mu$.
      \For{$i = 0$ to $M$}
      \State Collect rollouts and store them in their own on-policy training buffer $\mathcal{R}_i$ for each agent.
      \State Update the soft self-imitation training batches $\mathcal{D}_i$ for each agent.
      \State Compute advantage estimates $\hat{A}_i$, $i=\{1, 2, \dots, N\}$.
      \State Estimate the distance between current trajectories $\tau_i$ and highly-rewarded trajectories in soft self-imitation replay buffer $\mathcal{D}_i$ for each agent.
      \State Estimate the gradient $\nabla_{\theta}\tilde{L}_i$ in Eq.~\eqref{equ:nabla_P_I_sigma},
      $i=\{1, 2, \dots, N\}$.
      \State Perform policy improvement step:  
      $\theta_{i+\frac{1}{2}} \leftarrow \theta_i + \alpha\nabla_{\theta}\tilde{L}_i$ for each agent.
      \State Estimate $g$ and $H_{k}^{i}$.
      \State Perform policy exploration step by updating policy parameters according to Eq.~\eqref{equ:policy_explore_appro}.
      \EndFor
    \end{algorithmic}
  \end{algorithm}
\end{appendices}


\bibliography{reference}


\end{document}